\documentclass{article}

    \usepackage[final]{neurips_2020}

\usepackage[final]{neurips_2020}

\usepackage[utf8]{inputenc} 
\usepackage[T1]{fontenc}    
\usepackage{hyperref}       
\usepackage{url}            
\usepackage{booktabs}       
\usepackage{amsfonts}       
\usepackage{nicefrac}       
\usepackage{microtype}      
\usepackage{amsmath}        
\usepackage{todonotes}      
\usepackage{listings}
\usepackage{xcolor}
\definecolor{lg}{gray}{0.95}
\newcommand{\xpdone}[1]{
}

\usepackage{soul}

\title{A Framework For Contrastive Self-Supervised Learning And Designing A New Approach}

\author{%
  William Falcon\\
  New York University, NY\\
  Lightning Labs, NY \\
  \texttt{waf251@nyu.edu} \\
   \And
   Kyunghyun Cho \\
   New York University, NY\\
  \texttt{kc119@nyu.edu} \\
}
\begin{document}

\maketitle

\begin{abstract}
    Contrastive self-supervised learning (CSL) is an approach to learn useful representations by solving a pretext task which selects and compares anchor, negative and positive (APN) features from an unlabeled dataset. We present a conceptual framework which characterizes CSL approaches in five aspects (1) data augmentation pipeline, (2) encoder selection, (3) representation extraction, (4) similarity measure, and (5) loss function. We analyze three leading CSL approaches--AMDIM, CPC and SimCLR--, and show that despite different motivations, they are special cases under this framework. We show the utility of our framework by designing \textbf{Y}et \textbf{A}nother DIM (\textbf{YADIM}) which achieves competitive results on CIFAR-10, STL-10 and ImageNet, and is more robust to the choice of encoder and the representation extraction strategy. To support ongoing CSL research, we release the PyTorch implementation of this conceptual framework along with standardized implementations of AMDIM, CPC (V2), SimCLR, BYOL, Moco (V2) and YADIM.
\end{abstract}

\section{Introduction}
A goal of self-supervised learning is to learn to extract representations of an input using a large amount of unlabelled data. This representation is used to solve downstream tasks which often have only a few labelled instances. Self-supervised learning achieves this goal by solving a \textit{pretext task} which creates different types of supervision signals from unlabelled data based on careful inspection of underlying regularities in the data. In computer vision, there has been a stream of novel pretext tasks proposed over the past few years, including colorization~\cite{zhang2016colorful}, patch relatedness~\cite{doersch2015unsupervised,isola2015learning}, transformation prediction~\cite{agrawal2015learning,gidaris2018unsupervised}, in-painting~\cite{pathak2016context} and self-supervised jigsaw puzzle~\cite{noroozi2016unsupervised}. Recently, in computer vision, contrastive methods have achieved state-of-the-art results on ImageNet~\cite{hjelm2018learning,henaff2019data,chen2020simple,tian2019contrastive,he2019momentum}. In natural language processing, BERT~\cite{devlin2018bert} has become the {\it de facto} standard in low-resource text classification~\cite{liu2019roberta}.\footnote{
    see \url{https://gluebenchmark.com/leaderboard}.
} 
BERT is trained to predict (artificially) missing words given surrounding context as a pretext task. 

\begin{figure}[t]
  \centering
    \includegraphics[width=0.7\textwidth]{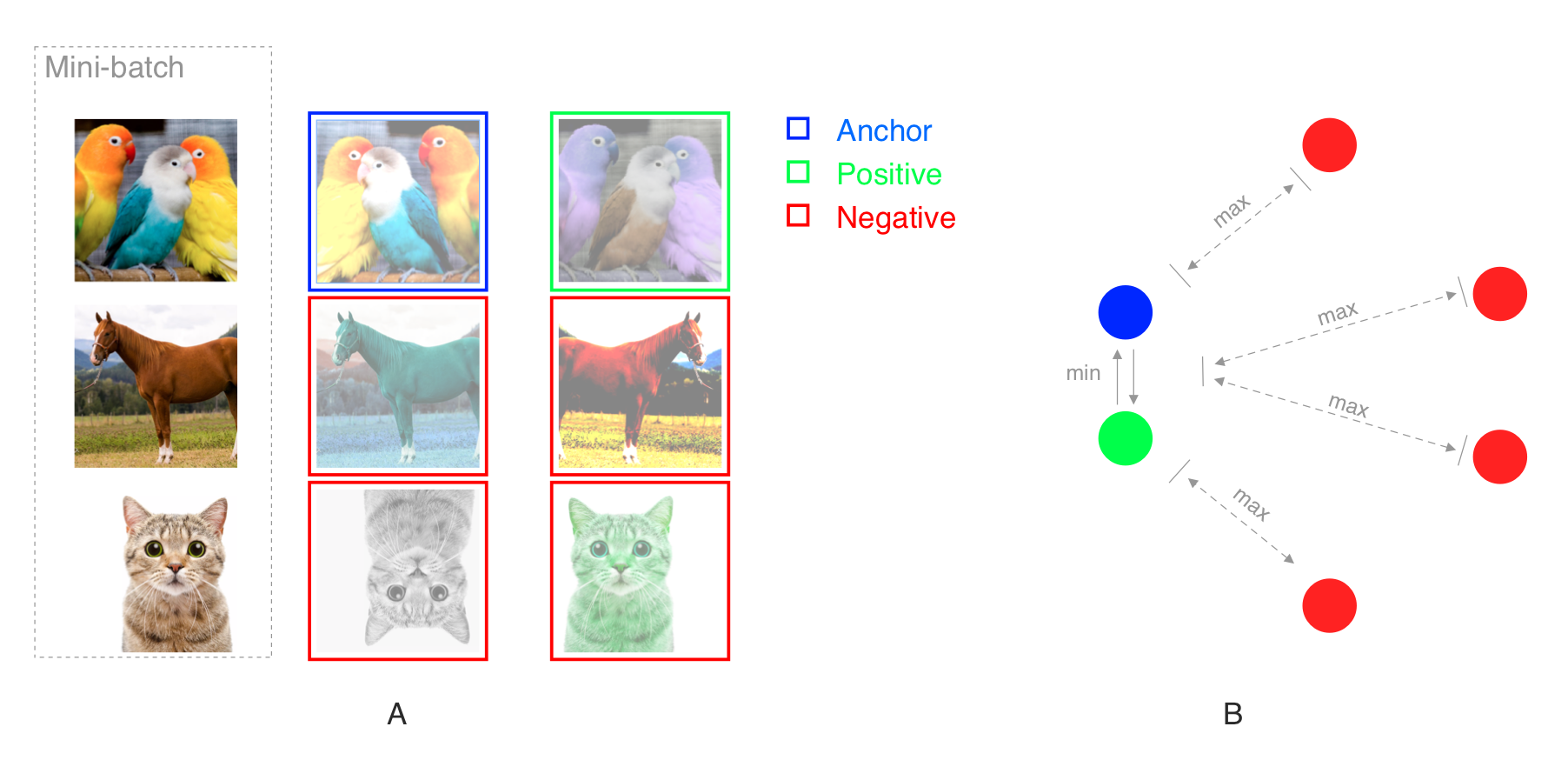}
    \caption{\textbf{(A)} The contrastive self-supervised pretext task consists of finding ways to generate anchor, positive and negative features used as training signals for unlabeled datasets. \textbf{(B)} The positive and anchor pairs are pushed close together, and away from the negative features.}
    \label{fig:ssl}
\end{figure}

To understand the differences between the CSL approaches in computer vision, we formulate a framework which characterizes CSL in five parts; (1) an encoder, (2) data augmentation pipeline, (3) a representation extraction, (4) a similarity measure and (5) a loss function. We use our framework to analyze three leading self-supervised learning approaches from which recent CSL approaches~\cite{chen2020simple,tian2019contrastive,he2019momentum} build on; augmented multi-scale deep information maximization (AMDIM; \cite{bachman2019learning,hjelm2018learning}), contrastive predictive coding (CPC; \cite{henaff2019data,oord2018representation}) and a simple framework for contrastive learning of visual representations (SimCLR; \cite{chen2020simple}). AMDIM was designed to maximize the mutual information (MI) between representations obtained from a single image while minimizing the MI between representations obtained from two separate images. CPC, on the other hand, extracts representations which can tell whether they come from the same or a different image. SimCLR, using the same ideas of AMDIM, maximizes the similarity between representations obtained from a single image, while minimizing the similarity between representations obtained from other images. 

Our analysis finds that despite different motivations behind these approaches, existing CSL algorithms are only slightly different from one another. We demonstrate the usefulness of the proposed framework by formulating a new CSL variant that merges the data processing pipelines of AMDIM and CPC. Our resulting approach, \textit{YADIM}, produces comparable results on downstream tasks including CIFAR-10, STL-10, and ImageNet, while improving the robustness to the choice of the encoder and sampling task. These results suggest that this framework is capable of generating CSL variant that perform just as well as the three leading approaches; AMDIM, CPC and SimCLR.

\section{Contrastive self-supervised learning}

Contrastive self-supervised learning (CSL) has been recently found successful for semi-supervised learning (SSL)~\cite{hjelm2018learning,henaff2019data,chen2020simple,tian2019contrastive,he2019momentum}. In CSL, the goal is to generate representations of instances such that similar instances are near each other and far from dissimilar ones. In supervised learning, associative labels determine the similarities among instances. Without labels however, we must develop methods to exploit similarities implicitly embedded within instances. CSL does this by generating anchor, positive, and negative samples from an unlabeled dataset.

Let $D = \{x_1, x_2, ..., x_N \}$ be an unlabeled dataset of size $N$. CSL builds on the assumption that each instance defines and belongs to its own class~\cite{dosovitskiy2014discriminative}. This assumption implies that we have $N$ classes. To create samples that belong to the same class we generate two features $(v^a, v^+)$ from the same example $x \in D$. We refer to $v^a$ as an \textbf{\textit{anchor}} feature and $v^+$ as a \textbf{\textit{positive}} feature. To create an example from a different class, we generate a feature $v^-$ from a different example $x'$. We call $v^-$ a \textit{\textbf{negative}} feature. Depending on the task, such a feature can be a vector $v \in \mathbb{R}^n$  or a multi-dimensional tensor $v \in \mathbb{R}^{n \times ... \times m}$. We propose the following five-part framework which lets us easily characterize existing CSL approaches.

\paragraph{(1) Data Augmentation Pipeline} 

The goal of the data augmentation pipeline is to generate anchor, positive and negative (APN) features to be used in contrastive learning. Let $a_n$ define a stochastic input augmentation process such as random flip and random channel drop. Then, $A = ( a_1, \ldots, a_N )$ defines a pipeline that applies these augmentations sequentially. We can apply $A$ to $x$ to generate a new sample $v^i$ which preserves the same underlying semantics as $x$. This strategy gives us a way to generate multiple samples of the same class defined by the example $x$ which we can use as a supervisory signal.

To generate the anchor and positive features we can take many approaches. One way to generate $v^a$ and $v^+$ is to sample two subsets of vectors from the same feature, $v^a, v^+ \subseteq v_x$. A second way is to apply $A$ to the same input twice, $v^a \sim A(x), v^+ \sim A(x)$, which produces two distinct sets of features due to the stochastic nature of $A$. The negative feature, $v^- \sim A(x')$ is sampled via the same process but taken from a different sample $x'$.

\paragraph{(2) Encoder}

Let $f_{\theta}$ define an encoder parameterized with $\theta$. This encoder can be any function approximator such as a fully-connected or convolutional neural network (CNN)~\cite{lecun2004learning,lecun1990handwritten,krizhevsky2012ImageNet}.
The encoder maps an input $v$ into a set of vectors $r$ which we call the \textit{representation} of $x$. When $x$ is an image with $s$ input channels, width $w$, and height $h$, then $f$ performs the following mapping $f_{\theta}:\mathbb{R}^{s \times w \times h} \to  \mathbb{R}^{k \times c}$. In other words the encoder returns $k$ $c$-dimensional feature vectors as the representation of the input. When the encoder is a convolutional network, $r$ is a set of vectors from a feature map $m$ where $m \in \mathbb{R}^{s \times w \times h}$.

\paragraph{(3) Representation extraction}

For contrastive learning we need to extract representations that can be compared against one another. Let $r^+ = f_{\theta}(v^+)$ be the \textit{positive} representation, $r^a = f_{\theta}(v^a)$ the \textit{anchor} representation and $r^- = f_{\theta}(v^-)$ the negative representation. Representations are extracted from an encoder or a sequence of encoders applied to $v^\cdot$. There are many ways to perform the representation extraction; one way is to generate a single $d$-dimensional vector as the final output of the encoder for each representation $r^\cdot \in \mathbb{R}^d$. Another way is to output a matrix for each representation $r^\cdot \in \mathbb{R}^{n \times k}$ and compare a subset of $r^a$ against another subset of $r^-$ to generate multiple negative scores.

\paragraph{(4) Similarity measure} 

Let $\Phi(r_a, r_b)$ measure the similarity between two representations, $r_a$ and $r_b$. This function outputs a scalar score $s$ which measures the similarity between $r_a$ and $r_b$. Examples of similarity measures are the dot product, cosine similarity, or bi-linear transformations such as $s = r_a^\top \cdot W r_b$, in which case $\Phi$ has its own parameter $W$.

\paragraph{(5) Loss Function}  

Refer to $s^+ = \Phi(r^a, r^+)$ as the positive score and to $s^- = \Phi(r^a, r^-)$ as the negative score. We define a loss function as the combination of the positive and negative scores to reflect the progress of learning. Minimizing this loss function corresponds to maximizing the positive score and minimizing the negative scores. 

Widely used loss functions include the \textit{negative contrastive estimation} (NCE) loss~\cite{mnih2013learning}, the \textit{triplet} loss~\cite{schroff2015facenet} and InfoNCE~\cite{henaff2019data}.

\subsection{Special Case 1: Augmented Multiscale Deep InfoMax (AMDIM)} \label{sec: amdim}

The first example of CSL that we describe under the proposed framework is AMDIM by Bachman~et~al.~\cite{bachman2019learning}, which was recently proposed for self-supervised learning of image representations. The motivation behind AMDIM is to maximize mutual information between features extracted from intermediate layers of a CNN, generated from two views of the same image. AMDIM operates on a dataset of $d$-channel images with width $w$ and height $h$. Let $D = \{x_1, x_2, ..., x_n \}$ be a collection of images $x \in \mathbb{R}^{w \times h \times d}$.

The implementation details of AMDIM can be found in Appendix~\ref{ap:amdim}.

\paragraph{(1) Data Augmentation Pipeline}

Let $x$ be an example from $D$, and $a_n$ a stochastic image augmentation stage. The augmentation pipeline for $x$ in AMDIM consists of five stages; random flip, image jitter, color jitter, random gray scale and normalization of mean and standard deviation. We refer the readers to \cite{bachman2019learning} for the details of each. AMDIM generates the \textbf{\textit{positive}} $v^+$ and \textbf{\textit{anchor}} $v^a$ features by applying $A$ twice to the same input $x$, $v^a \sim A(x)$ and $v^+ \sim A(x)$. The \textbf{\textit{negative}} $v^-$ feature is generated by applying $A$ to a different input $v^- \sim A(x')$.

\paragraph{(2) Encoder}

The encoder, $f_{\theta}$, in AMDIM is based on a residual network (ResNet)~\cite{he2016deep} with three design considerations. The first ensures that the encoder design minimizes information shared between two feature vectors from the same image. To achieve this, Bachman~et~al. minimize the overlap between receptive fields and do not use batch normalization~\cite{ioffe2015batch}. Second, the encoder does not use padding in order to avoid learning artifacts introduced by padding. Third, the number of channels in this modified ResNet is 5 times more than the number of channels found in ResNet-34 which is the most similar commonly-used architecture. Specifically, ResNet-34 has 64, 128, 256 and 512 feature maps, whereas the AMDIM ResNet has 320, 640, 1280 and 2560 feature maps, respectively. AMDIM was tested to work with this particular variant of a ResNet.

\paragraph{(3) Representation extraction}

AMDIM draws $(r^a, r^+, r^-)$ triplets from feature maps extracted at different scales. Here we offer a brief summary of this task. Let $M = \{ m_1, m_2, ..., m_l\} = f_{\theta}(x)$ define the set of all feature maps generated by an encoder from each layer $l$ when applied to an input $x$. Let $M^a = f_{\theta}(v^a), M^+ = f_{\theta}(v^+)$ and $M^- = f_{\theta}(v^-)$. 

Let $m_j$ and $m_k$ be two feature maps at layers $j$ and $k$. The anchor representation $r^a \sim m^a_j$ is taken from $m^a_j \subset M^a$. Multiple positive representations $R^+$ are used, where $R = m^+_{k}$ are taken from the $m^+_{k} \subset M^+$. To create the negative samples we use all the $m^-_{k}$ generated from every other example. These samples define the set of negative representations, $R^- = \underset{m_i \in M_{x'}} \bigcup  m^-_{i}$. In other words, AMDIM requires an encoder that returns representations of the inputs at multiple scales.

\paragraph{(4) Similarity Measure}

AMDIM uses a dot product between two vector representations with $c$ channels, $\Phi(a, b) = a \cdot b$ to measure their similarity, where $a \in \mathbb{R}^{c}, b \in \mathbb{R}^{c}$ and $\Phi(a, b) \in \mathbb{R}$. This means that the dimensions of intermediate representations generated by the encoder must be the same.

\paragraph{(5) Loss Function}

AMDIM uses an NCE loss $\mathcal{N}_{\theta}(r^a, R^+, R^-)$ given a representations triplet~\cite{mnih2013learning}:
\begin{align}
    \label{eq:nce}
    \mathcal{N}_{\theta}(r^a, R^+, R^-) = - \log \frac{\sum_{r^+_i \in R^+}{\exp({\Phi(r^a, r^+_i)})}}
    {\sum_{r^-_i \in R^-} \exp(\Phi(r^a, r_i^-))}.
\end{align}

AMDIM minimizes the NCE loss above from the following combinations of the final three feature maps obtained from the encoder:
\begin{align}
    \label{eq:amdim}
    \mathcal{L}_{\text{AMDIM}} = - \frac{1}{3} \left[\mathcal{N}_{\theta}(r^a_j, R^+_{k-1}, R^-_{k-1}) + \mathcal{N}_{\theta}(r^a_j, R^+_{k-2}, R^-_{k-2}) + \mathcal{N}_{\theta}(r^a_{j-1}, R^+_{k-1}, R^-_{k-1}) \right],
\end{align}
requiring an encoder that outputs at least three feature maps.

\subsection{Special Case 2: Contrastive Predictive Coding (CPC)} 
\label{sec: cpc}

The second example of CSL we describe under the proposed framework is contrastive predictive coding (CPC) by Henaff~et~al~\cite{henaff2019data}. CPC learns to extract representation by making the feature vector of a patch from an image predictive of another patch spatially {\it above} the original patch. CPC differs from AMDIM in the encoder design, data augmentation pipeline, sampling task and loss function.

The implementation details can be found in Appendix~\ref{ap:cpc}.

\paragraph{(1) Data Augmentation Pipeline}   

The CPC data augmentation pipeline is the same as that of AMDIM with the following new transformation stage $a'$ appended at the end. This stage $a'$  breaks an input image $x$ with width $w$, height $h$ and channels $d$, into a list of $p$-many $(q \times q)$ overlapping patches with $d$ channels each. This transformation maps $x: \mathbb{R}^{w \times h \times d} \to \mathbb{R}^{p \times q \times q \times d}$. For example, for a $256 \times 256$ image, patch size $q=64$ with a 32-pixel overlap, $x: \mathbb{R}^{256 \times 256 \times 3} \to \mathbb{R}^{7 \times 7 \times 64 \times 64 \times 3}$.

Define $v \sim A(x)$ as the set of features for image $x$, and $v'$ the set of features for image $x'$.

\paragraph{(2) Encoder}  

The encoder $f_{\theta}$ in CPC is a modified version of the ResNet-101~\cite{he2016deep}. The ResNet-101 has four stacks of residual blocks. Each residual block has three convolutional layers of which the second one, \textit{the bottleneck layer}, produces a fewer feature maps. CPC modifies the third stack in three ways; 1) they double the number of residual blocks from 23 to 46, 2) they double the bottleneck layer dimension from 256 to 512, and 3) they increase the number of feature maps from 1024 to 4096. CPC replaces all the batch normalization~\cite{ioffe2015batch} layers with layer normalization~\cite{ba2016layer} to minimize information sharing between two feature vectors from the same image.

\paragraph{(3) Representation extraction}

Let $H = f_{\theta}(v)$ be the output of the encoder, where $v \sim A(x)$ and $H \in \mathbb{R}^{c \times h \times w}$. CPC frames representation extraction as predicting the feature vectors $k$ units spatially below at $H_{i + k, j}$ based on the feature vectors in the neighbourhood of $H_{i,j}$. 

For example, the vector at row $0$ is used to predict the vectors at all the rows $1, 2, ..., h$. This prediction task first applies masked convolution to $H$ via a \textit{context encoder} $g_{\psi}$ to generate $C = g_{\psi}(H)$,  where $C \in \mathbb{R}^{c \times h \times w}$. Each $c_{i, j}$ summarizes the context around every $H_{i, j}$. For each $c_{i ,j}$, the target is then chosen from each row whose index is $k > i$. Prediction $\hat{H}_{i+k, j} = W_k c_{i, j}$  is finally generated using a prediction matrix $W_k$. 

This prediction $r^a = \hat{H}_{i+k, j}$ is the \textit{\textbf{anchor representation}}, the target to predict $r^+ = H_{i+k, j}$ the \textit{\textbf{positive representation}}, and all the other $H \backslash \{ h_{i, j} \}$ serve as the \textit{\textbf{negative}} representations $R^-$. This task can be viewed as correctly picking $\hat{H}_{i+k, j}$ from the set which has the real $H_{i+k, j}$ and distractors $H_{i+k, j}', H_{i+(k+1), j}', \ldots, H_{i+k+n, j}'$.

\paragraph{(4) Similarity Measure}

CPC uses a dot product of $\hat{H}_{i+k, j}$ and $H_{i+k, j}$.

\paragraph{(5) Loss Function}

Just like AMDIM, CPC uses the NCE loss:
\begin{align}
    \label{eq:cpc}
    \mathcal{L}_{\theta}(r^a, r^+, R^-)=
    - \log \frac{\exp({\Phi(r^a, r^+_i)})}
    {\exp({\Phi(r^a, r^+_i)}) + \sum_{r^-_i \in R^-} \exp(\Phi(r^a, r_i^-))}.
\end{align}
This loss drives the anchor and positive samples together while driving the anchor and negative samples apart.

\subsection{Special Case 3: A Simple Framework for Contrastive Learning of Visual Representations (SimCLR)} 
\label{sec: simclr}

The third example of CSL we describe under the proposed framework is a simple framework for contrastive learning of visual representations (SimCLR) by Chen~et~al~\cite{chen2020simple}. SimCLR extracts representation by maximizing the similarity between representations extracted from two views of the same image, just like AMDIM does. SimCLR is similar to AMDIM with a series of a few minor tweaks. First, it uses a non-customized, generic ResNet. Second, it uses a modified data augmentation pipeline. Third, it adds a parametrized similarity measure using a projection head. Finally, it adds a scaling coefficient ($\tau$) to the NCE loss.

The implementation details of SimCLR can be found in Appendix~\ref{ap:simclr}.

\paragraph{(1) Data Augmentation Pipeline}

The augmentation pipeline of SimCLR follows the same ideas introduced by AMDIM. This pipeline applies a stochastic augmentation pipeline twice, to the same input $v^a \sim A(x), v^+ \sim A(x)$. The data augmentations consist of random resize and crop, random horizontal flip, color jitter, random gray scale and Gaussian Blur. We refer the reader to \cite{chen2020simple} for the details of each.

\paragraph{(2) Encoder}  

The encoder $f_{\theta}$ in SimCLR is a ResNet of varying width and depth. The ResNets in SimCLR use batch normalization.

\paragraph{(3) Representation extraction}

Let $r = f_{\theta}(v)$ be the output of an encoder, where $r \in \mathbb{R}^{c \times h \times w}$. We obtain vector by reshaping $r: \mathbb{R}^{c \times h \times w} \rightarrow \mathbb{R}^{c \cdot h \cdot w}$. The representation $r^a = f_{\theta}(v^a)$ is the \textit{\textbf{anchor representation}}, $r^+ = h^+ = f_{\theta}(v^+)$ the \textit{\textbf{positive representation}} and $R^-$ the set of negative representations generated from all the other samples $x'$.

\paragraph{(4) Similarity Measure}

SimCLR uses a projection head $z = f_{\phi}$ to map the representation vector from the encoder to another vector space, i.e., $f_{\phi}: \mathbb{R}^c \rightarrow \mathbb{R}^c$. The cosine similarity between the $(z_i, z_j)$ pair is used as a similarity score. The composition of the projection and cosine similarity can be viewed as a parametrized similarity measure.

\paragraph{(5) Loss Function}

SimCLR uses the NCE loss with a temperature $\tau \in \mathbb{R}$ to adjust the scaling of the similarity scores
\begin{align}
    \label{eq:simclr}
    \mathcal{L}_{\theta}(z^a, z^+, Z^-)=
    - \log \frac{\exp({\Phi(z^a, z^+)/\tau})}
    {\sum_{z^-_i \in Z^-} \exp(\Phi(z^a, z_i^-)/\tau)}.
\end{align}
This loss drives the anchor and positive samples together while driving the anchor and negative samples apart.

\section{From CPC and AMDIM to YADIM}

In this section we show the usefulness of our conceptual framework by creating Yet Another variant of DIM, called \textit{YADIM}, which combines the ideas from both CPC and AMDIM. We first introduce the general design principles behind YADIM. For the components that differ between CPC and AMDIM, we perform ablations to determine which one to use. We end the section by summarizing the final YADIM design.

\subsection{YADIM: Yet Another DIM}

When viewed under our framework, CPC and AMDIM are more similar than expected from their different, underlying motivations. However, they do differ on the particular encoder design, data augmentation pipeline, representation extraction, and the similarity measure, while using the same loss. We formulate YADIM by empirically assessing the impact of each of these subtle design differences on the performance of both AMDIM and CPC.
When any particular design choice does not have much impact on the performance,  we choose the (simplified) union of both. We focus on CPC and AMDIM, as we found SimCLR to be a minor variant of AMDIM. Below, we present the general YADIM formulation.

\paragraph{(1) Data Augmentation Pipeline}

Recall the data augmentation pipeline of AMDIM which consists of five stages;  
random flip, 
image jitter, 
color jitter, 
random gray scale and 
z-normalization. CPC appends to this pipeline by AMDIM a transform which breaks an input image $x$ with width $w$, height $h$ and channels $d$, into a list of $p$-many $(q \times q)$ overlapping patches with $d$ channels each. This transform maps $x: \mathbb{R}^{w \times h \times d} \to \mathbb{R}^{p \times q \times q \times d}$

The independent nature of each pipeline lends itself to a natural joint formulation that is the union of the CPC and AMDIM pipelines. The YADIM pipeline applies all six transforms to an input in sequence to generate two version of the same input, $v^a \sim A(x)$ and $v^+ \sim A(x)$, and uses it to generate the negative sample from a different input $v^- \sim A(x')$.

\paragraph{(2) Encoder}

CPC and AMDIM both use customized CNN encoders. For YADIM we use the same encoder used by AMDIM.

\paragraph{(3) Representation extraction}

AMDIM compares the triplets of representations generated by the encoder at different spatial scales, while CPC uses a context encoder to predict a representation spatially lower in a feature map. These two tasks make specialized and unique domain assumptions which makes them difficult to merge. Instead, we test variations of the AMDIM sampling task. For YADIM, we eventually choose the task with the fewest assumptions while achieving the near-best performance.

\paragraph{(4) Similarity measure}

AMDIM uses a dot product $\phi(a, b) = a \cdot b$, while CPC uses a parametrized dot product between the representation $\hat{H}_{i+k, j} = W_k c_{i, j}$ and the target representation $H_{i+k, j}$. For YADIM, we resort to using dot product, because the encoder can subsume linear transformation in the parametrized measured used by CPC, if necessary.

\paragraph{(5) Loss Function}

YADIM uses the NCE loss which was used by both CPC and AMDIM.

\subsection{Systematic investigation of CPC and AMDIM design differences}

For the two main elements of CPC and AMDIM that differ, the encoder and representation extraction, we perform ablative experiments on each and investigate how each alternative decision affects CPC, AMDIM and YADIM.

\paragraph{Encoder Architecture}

CPC uses a modified ResNet-101 with wider layers, while AMDIM uses a modified ResNet-34. In this ablation, we replace the encoder in each approach with 9 standard ResNet architectures taken from the torchvision package.\footnote{
    \url{https://pytorch.org/docs/stable/torchvision/index.html}
}
For the baseline YADIM, we use the wide ResNet-34 which is the encoder from AMDIM. Table~\ref{tab: ResNet-ablations} shows the sensitivity of each approach to the choice of encoder.

\begin{table}[t]
    \begin{minipage}{0.40\textwidth}
      \caption{\textbf{Encoder robustness on CIFAR-10.} (higher is better, \textbf{bold} results are the highest per ResNet). All major ResNet architectures found in PyTorch~\cite{paszke2019pytorch} are trained with either AMDIM, CPC or proposed YADIM. We pretrain each network on CIFAR-10 without labels and train an MLP with 1,024 hidden units on CIFAR-10 with all labels on top of the pretrained and frozen encoder. YADIM's performance remains stable across different choice of encoders, as does CPC while AMDIM performance suffers.
      }
      \label{tab: ResNet-ablations}
    \end{minipage}
    \begin{minipage}{0.60\textwidth}
      \centering
      \begin{tabular}{l|ccc}
        Network & AMDIM & CPC & YADIM\\
        \midrule
        Reported & 93.10 \xpdone{} & - &  -\\ 
        Our implementation & 92.00 \xpdone{(amdim-paper-cf10-1)} & {84.52} & 90.33  \\   
        \hline
        ResNet 18  & 63.25 \xpdone{(amdim-r18-1)} & {83.14} & \textbf{85.51} \xpdone{(iclr-patches-ResNets1-clf-2)}\\
        ResNet 34  & 57.50 \xpdone{(amdim-r34-1)} & {83.91} & \textbf{84.57} \xpdone{(iclr-patches-ResNets1-clf-5)}\\
        ResNet 50  & 61.43 \xpdone{(amdim-r50-1)} & {83.76} & \textbf{85.83} \xpdone{(iclr-patches-ResNets1-clf-6)}\\
        ResNet 101  & 58.91 \xpdone{(amdim-r101-1)} & {79.18} & \textbf{83.09} \xpdone{(amdim-ddt-6-ResNet-1-clf-9)}\\
        ResNet 152  & 53.40 \xpdone{(amdim-r152-1)} & {84.21} & \textbf{85.03} \xpdone{(iclr-patches-ResNets1-clf-12)}\\
        ResNet 50 (32 x4d)  & 64.08 \xpdone{(amdim-r50-32-x4d-1)} & {85.08} & \textbf{86.69} \xpdone{(iclr-patches-ResNets1-clf-15)}\\
        ResNet 101 (32 x8d)  & 59.37 \xpdone{(amdim-r101-32-x8d-1)} & {83.65} & \textbf{86.62} \xpdone{(iclr-patches-ResNets1-19)}\\
        Wide-ResNet 50  & 59.37 \xpdone{(mdim-wr50-2-1)} & {85.09} & \textbf{85.65} \xpdone{(iclr-patches-ResNets1-clf-21)}\\
        Wide-ResNet 101  & 60.07 \xpdone{(mdim-wr101-2-1)} & {84.15} & \textbf{85.60} \xpdone{(iclr-patches-ResNets1-clf-24)}\\
      \end{tabular}
    \end{minipage}
\end{table}

The first column demonstrates the drop in performance with AMDIM performance when using standard ResNets as the encoder. The second column shows that CPC does not suffer from the same drop in performance when using a standard ResNet. In the final column, we observe that YADIM is less sensitive to the choice of encoder. These results suggest that the choice of encoder in YADIM and CPC is less important than it was with AMDIM. Although It has recently been noted by~\cite{chen2020simple} that the network width has a significant impact on the performance, we do not observe such a dramatic difference in performance at least with CPC and YADIM. Furthermore, in AMDIM, the widest network (Wide-ResNet 101) performs worse than the smallest, non-wide network (ResNet-18).





\paragraph{Representation extraction}

AMDIM compares feature maps at different stages of an encoder, while CPC simply uses the last feature map. To evaluate the impact of each comparison approach, we evaluate AMDIM with a similar strategy and try two others that we design.

Let $(j:k)$ denote comparing feature maps, $m^a_j$ and $m^+_k$, and we use $j=-1$ to refer to the final feature map generated by the encoder and $j=-2$ second to the last. With this notation, AMDIM performs the comparison of $(-1:-2) + (-1:-3) + (-2:-2)$. To evaluate the sensitivity to the choice of feature map locations, we define five comparison strategies: (1) the last feature maps only $(-1: -1)$, (2) the AMDIM strategy $(-1:-2) + (-1:-3) + (-2:-2)$, (3) the last feature map to a random feature map $(-1:k\sim U(-1, -k))$, and (4) the feature maps at all levels separately $(-1:-1) + (-2:-2) + (-3:-3)$. In Table~\ref{tab: apn-ablations}, we see that AMDIM is highly sensitive to the particular representation extraction comparison, which is not the case with the proposed YADIM.

\begin{table}[t]
    \caption{\textbf{Robustness to the representation extraction approach on CIFAR-10:} Here we show the results of using a simple representation extraction for each approach. This task is to use only the last feature map from the encoder for contrastive learning. The success of AMDIM depends heavily on the particular task, whereas YADIM does not.
  }
  \label{tab: apn-ablations}
  \center
  \begin{tabular}{l|ccc}
    Representation extraction & AMDIM & YADIM \\
    \toprule
    Reported  & 90.00 & 91.98 
    \xpdone{(iclr-patches-cf10-111517-clf0)}\\
    \midrule
    AMDIM task & 90.00 & 92.40 \\
    Last only & 64.00 & 91.98 \\
    Last + random & 79.00 & 92.20 \\
    Same level  & 82.00 & 92.22 \\
  \end{tabular}
\end{table}

\subsection{Final formulation of YADIM}

Given the results of the ablations above, we finalize the design of the proposed YADIM in this section.

\paragraph{(1) Data Augmentation Pipeline}

We define a data augmentation pipeline for YADIM as the union of the CPC and AMDIM pipelines. This new pipeline applies all six transforms sequentially to an input twice to generate two version of the same input $v^a \sim A(x), v^+ \sim A(x)$. The same pipeline generates the negative sample from a different input $v^- \sim A(x')$.

\paragraph{(2) Encoder}

We use the wide ResNet-34 from AMDIM, although the choice of any other encoder would not have a significant impact the final performance.

\paragraph{(3) Representation Extraction}

YADIM compares the triplets from the last feature maps generated by the encoder $(r^a_{-1}, r^+_{-1}, R^-_{-1})$, unlike AMDIM.

\paragraph{(4) Similarity measure}

YADIM uses a dot product $\phi(a, b) = a \cdot b$ without any extra parameter, unlike CPC.

\section{Experimental setup}

\paragraph{Datasets}

We use four standard image classification datasets to thoroughly evaluate the proposed approach together with AMDIM and CPC:
\begin{itemize}
    \item  \textbf{CIFAR-10}~\cite{krizhevsky2009learning} consists of 32x32 images categorized into 10 classes. It has 50,000 training images and 10,000 test images. We use a 45,000/5,000 training/validation split. We report results on the 10,000 test images.
    
    \item \textbf{STL-10}~\cite{coates2011analysis} consists of 96x96 images categorized into 10 classes. We downsample each image to be 64x64 following Bachman~et~al.~\cite{bachman2019learning}. 
    The training set has 100,000 unlabeled and 5,000 labeled training examples. The test set has 8,000 labeled examples. We train our model and baselines using the unlabeled training examples, while using 500 labeled ones for validation.
    
    \item \textbf{ImageNet (ILSCRV-12)}~\cite{deng2009ImageNet} has images of varying sizes categorized into 1,000 classes. We downsample each image to be 128x128 following the protocol from~\cite{bachman2019learning}. The dataset has approximately 1.3 million images in the training set and 50,000 images in the official validation set. We make our own 50,000 image validation set from the training set and use the official validation set as the test set.
\end{itemize}

\paragraph{Optimization and Hyperparameters} 

We use Adam~\cite{kingma2014adam} and search for the optimal learning rate via grid search based on the loss function computed on the unlabeled training examples. We fix the other hyperparameters of Adam to the same values used in \cite{bachman2019learning}. For SimCLR, we use LARS~\cite{you2017large}, as originally used by the authors, with a learning rate of $10^{-6}$.

\paragraph{Finetuning}
\label{section:ft_protocol}

In \cite{kolesnikov2019revisiting}, the authors conducted extensive study on how to evaluate self-supervised learning algorithms for image representation. We thus follow their evaluation protocol in our experiments: 1) choose a dataset, 2) drop all labels, 3) pretrain the encoder on these unlabeled examples, 4) freeze the encoder, and 5) train a separate neural network on top of this frozen encoder using a subset of labeled examples.
%

We use a multi-layer perceptron (MLP) with a single hidden layer consisting of 1,024 ReLU ~\cite{nair2010rectified} units. Our results differ slightly from those reported because different CSL approaches use different variants of MLP or ResNet.
We however find it more informative for comparison to use the same MLP across all CSL approaches.

\paragraph{Compute Infrastructure}

We use multiple NVIDIA V100 GPUs with 32G memory each, for each experiment. We use two V100 GPUs for up to 24 hours on CIFAR-10. We use eight V100 GPUs for up to 72 hours on STL-10. We use 96 V100 GPUs for up to 72 hours on ImageNet when AMDIM is used and 32 V100 GPUs for 21+ days when CPC was used. YADIM trained for 14 days using 256 V100 GPUs to cope with the increased memory requirement incurred by the use of both non-overlapping patches and double applications of data augmentation.

\subsection{Implementation}

The CSL approaches often use different evaluation protocols and do not re-implement individual approaches for rigorous comparison~\cite{henaff2019data,tian2019contrastive,hjelm2018learning,chen2020simple}. To avoid any inconsistency arising from different implementations and protocols, we re-implement AMDIM, CPC, SimCLR, Moco and CMC using PyTorch Lightning.

\paragraph{PyTorch Lightning}

PyTorch Lightning (PL)~\cite{falcon2019pytorch} is a framework which decouples scientific components and engineering details in the code written for PyTorch~\cite{paszke2019pytorch}. PL enables our implementations of the CSL approaches to be hardware agnostic, more easily readable, and accessible to researchers with lower computational resources since it enables running the same code on arbitrary hardware. In addition, it allows us to use the exactly same dataset splits, same fine-tuning protocol, early-stopping criterion and transformation pipelines to ensure the consistency across various experimental settings.

\paragraph{Negative samples}

Through our experiments, we observe that CSL performance positively correlates with the number of negative samples. There are multiple ways to achieve this. One way is to build a memory bank that stores pre-computed representations for $k$ consecutive samples~\cite{tian2019contrastive,chen2020simple}. Another way is to share data across training processes, where all the batches across GPUs can be used for the denominator of  softmax~\cite{falcon2019pytorch,miller2017parlai}.

For our experiments we use distributed softmax via PyTorch Lightning, which splits a batch of samples across GPUs on the machine and aggregates them on a single machine for an effective large denominator for softmax. 

\paragraph{CSL reproducibility}

As will be evident from the results later, we have largely reproduced AMDIM and CPC which we attribute to the open-sourced code of AMDIM provided by its authors and helpful discussion with the authors of CPC, respectively. Due to the limitation on available computational resources, however, we were not able to perform thorough hyperparameter search with CPC and AMDIM, which prevented us from fully reproducing their reported results on ImageNet.

\section{Results and analysis}

In this section we verify our implementations. To our knowledge, this is the first comparison between AMDIM, CPC and SimCLR using the same standardized implementation and evaluation protocol.

\paragraph{Class Separability}

In this set of experiments, we test whether representations capture class separation without observing the actual class labels. To measure this, we apply the fine-tuning protocol we describe in \ref{section:ft_protocol}. For training a classifier, we use the all the labels of CIFAR-10, STL-10 and ImageNet.

Table~\ref{tab: amdim-ft_ImageNet} compares the class separability of the representations from AMDIM, CPC, SimCLR and YADIM. Our implementation of AMDIM achieves the performance close to that reported by Bachman~et~al. on CIFAR-10 and STL-10. On ImageNet, our implementation lags behind the latest reported accuracy but is still on par with the original accuracy from the earlier version of \cite{bachman2019learning}.

Our CPC implementation sets new state-of-the-art scores on STL-10 self-supervised pretrained models. On ImageNet our CPC implementation achieves close to the reported result, but due to computational constraints we cannot train the model fully to achieve the accuracy reported by Henaff~et~al.~\cite{henaff2019data}. Our SimCLR implementation is trained on a single V100 for 14 hours and achieves 87.60 accuracy which is reasonably close to the reported $93.70$. Finally, the proposed YADIM achieves comparable accuracy on CIFAR-10, SLT-10 and ImageNet to both AMDIM and CPC.

\begin{table}[t]
  \caption{\textbf{Class separability} Test accuracies are computed from finetuning a 1,024-unit MLP on top of a frozen, pretrained encoder. The encoder was trained on each dataset without labels. We underline the best accuracy per dataset inclusive of previously reported ones, and bold-face the best accuracy among our own implementations. (1,2) reported in the first and second versions of \cite{bachman2019learning}, respectively. Due to computational constraints we could not complete experiments using our own implementation of SimCLR on ImageNet. $\dagger$ our own implementation. 
  $\star$ reported in \cite{henaff2019data}.
  }  
  \label{tab: amdim-ft_ImageNet}
  
  \centering
  \begin{tabular}{l|lll}
    \toprule
    Method     & CIFAR-10 &  STL-10 & ImageNet\\
    \midrule
    AMDIM$^{(1)}$ & 93.10 & 93.80 &
    60.20  \xpdone{(amdim-ImageNet-1000-1-clf-1-cvpr-ddp-2-version-2)} \\   
    AMDIM$^{(2)}$ & \underline{93.10} & \underline{93.80} &
    68.10 \xpdone{(am-img-1000-1-clf-1)} \\  
    CPC$^\star$ & -- & -- & 64.03 \\   
    SimCLR$^\star$ & 93.70 & -- & 71.70 \\   
    \hline 
    AMDIM$^\dagger$  & \textbf{92.10}
    & \bf 91.50
    & \textbf{60.08}
    \xpdone{(amdim_ImageNet_1000_1_clf_1_cvpr_ddp_2)} 
    \\   
    CPC$^\dagger$ & 84.52  & 78.36 & 54.82 \\   
    SimCLR$^\dagger$ & 87.60 & -- & -- \\   
    YADIM & {91.30} & {92.15 } & 59.19
    \\
    \bottomrule
  \end{tabular}
\end{table}  

\section{Related work}  

Pretext tasks for self-supervised learning have been studied extensively over the past few years. In \cite{doersch2015unsupervised} the authors sample two neighboring patches within an image, then train a siamese network to predict the relative location of the patches. Isola~et~al.~\cite{isola2015learning} use a similar approach but instead predict whether patches were taken from nearby locations in the image or not. Pathak~et~al.~\cite{pathak2016context} attempt to learn representations by inpainting~\cite{bertalmio2000image}.
Noroozi~et~al.~\cite{noroozi2016unsupervised} train a CNN to solve jigsaw puzzles.
In a different approach,~\cite{zhang2016colorful} use the gray components of an image to predict the color components of the same image. More recently, Gidaris~et~al.~\cite{gidaris2018unsupervised} train CNNs to detect a 2d rotation applied to the input image.

Unlike the ones above which are specific to images, there have been a class of self-supervised learning algorithms that are less specific to images and are more generally applicable. They include  AMDIM~\cite{hjelm2018learning}, CPC~\cite{henaff2019data}, SimCLR~\cite{chen2020simple}, CMC~\cite{tian2019contrastive} and MOCO~\cite{chen2020mocov2,he2019momentum}. We refer to them as contrastive self-supervised learning (CSL). In this paper we attempt at providing a unified framework behind these CSL algorithms.

Dosovitskiy~et~al.~\cite{dosovitskiy2014discriminative} introduce the idea of generating multiple views of the same data by applying a stochastic augmentation pipeline to the same image multiple times. AMDIM pairs this idea with comparisons of feature maps generated from intermediate layers of an encoder. However, in this work we showed that the particular choice of which feature maps to compare is highly subjective and that performance deteriorates as the strategy changes. CPC on the other hand introduces the idea of generating positive pairs by taking patches within the same image and then predicts patches well-separated spatially. However, we observe that removing the context encoder of this task results in trivial solutions for both CPC and SimCLR.

CMC uses the same ideas from AMDIM but differs in two key aspects. First, the NCE loss is regularized by a discriminator~\cite{goodfellow2014generative}. Second, a memory bank is used to increase the number of negative samples, which leads to the increase in the memory requirement of the system. SimCLR differs from AMDIM in three key aspects. First, it adds random resize and crop to the data augmentation pipeline. Second, it parametrizes the similarity metric with a non-linear transformation of the representation followed by dot product. MOCO (v2) modifies the SimCLR projection head $g_{\theta}(x)$ and the data augmentation pipeline.

As Hjelm~et~al.~\cite{hjelm2018learning} demonstrated, CSL approaches outperform other approaches such as autoencoders \cite{rumelhalt1986learning,baldi1989neural} and generative adversarial networks (GAN) \cite{goodfellow2014generative}. Zhang~et~al.~\cite{zhang2016colorful} similarly showed that  representations learned by GANs and autoencoders do not transfer well to other tasks including image classification, although these models excel at image denoising and image synthesis.

\section{Conclusion} 

In this work we proposed a conceptual framework to more easily characterize various CSL approaches. We showed that AMDIM, CPC and SimCLR are special cases of our framework. We evaluated each key design choice of AMDIM and CPC, which are two representative CSL algorithms, and used our framework to construct a new approach to which we refer as YADIM (Yet another DIM). YADIM performs just as well as CPC and AMDIM on CIFAR-10, STL-10 and ImageNet but has two key advantages. First, it is robust to the choice of the encoder architecture. Second, it uses a simpler representation extraction strategy. 

By comparing the proposed YADIM against AMDIM and CPC, we have learned three lessons. First, the choice of encoder is not important as long as it is wide with many feature maps at each layer. Second, it is enough to use  a simple contrastive loss, consisting of noise contrastive loss with dot product without any extra parameter. Third, it is largely a strong data augmentation pipeline that leads to strong downstream task results, even with a simple representation extraction strategy or a simple encoder architecture. Furthermore, we find that SimCLR, CMC and MOCO
do not differ much from each other and from both CPC and AMDIM, and the design choices they make are easily interpreted under the proposed conceptual framework.

Finally, we release all the code for implementing these contrastive self-supervised learning approaches under PyTorch Lightning. We hope this release enables objective and clear comparison between all approaches and encourages researchers to push the frontier of contrastive self-supervised learning further.

\section*{Acknowledgement}

WF thanks Facebook AI Research, DeepMind and NSF for their support. In addition, thank you to Stephen Roller, Margaret Li, Shubho Sengupta, Ananya Harsh Jha, Cinjon Resnick, Tullie Murrell, Carl Doersch, Devon Hjelm, Eero Simoncelli and Yann LeCun for helpful discussions.
KC thanks support by CIFAR, NVIDIA, eBay, Google and Naver.

\bibliography{neurips_2020.bib}
\bibliographystyle{plain}

\medskip

\newpage
\appendix
\section{AMDIM bolts implementation code}\label{ap:amdim}

\subsection{Encoder}

\begin{lstlisting}
from pl_bolts.models.self_supervised.amdim import AMDIMEncoder
\end{lstlisting}

\subsection{Transforms}
\begin{lstlisting}
from pl_bolts.models.self_supervised.amdim import AMDIMEvalTransformsSTL10

dataset = STL10(transforms=AMDIMEvalTransformsSTL10())
\end{lstlisting}

\subsection{Representation Extraction}
\begin{lstlisting}
from pl_bolts.losses.self_supervised_learning import FeatureMapContrastiveTask

amdim_task = FeatureMapContrastiveTask('01, 02, 11')
\end{lstlisting}

\subsection{Loss}

\begin{lstlisting}
from pl_bolts.losses.self_supervised_learning import AmdimNCELoss

def contrastive_task_forward(M1, M2):
    (ma0, ma1, ma2) = M1
    (mb0, mb1, mb2) = M2
        
    01_loss = AmdimNCELoss(ma0, mb1) + AmdimNCELoss(mb0, ma1) 
    02_loss = AmdimNCELoss(ma0, mb2) + AmdimNCELoss(mb0, ma2) 
    11_loss = AmdimNCELoss(ma1, mb1) + AmdimNCELoss(mb1, ma1)
    return 01_loss + 02_loss + 11_loss
\end{lstlisting}

\subsection{Full AMDIM pseudocode}
The PyTorch Lightning implementation of AMDIM can be summarized by:

\begin{lstlisting}
def training_step(self, batch, batch_idx):
    x, _ = batch
    xa = augmentations(x)
    xb = augmentations(x)
    
    (ma0, ma1, ma2) = encoder(xa)
    (mb0, mb1, mb2) = encoder(xb)
    
    loss = contrastive_task(
            (ma0, ma1, ma2), 
            (mb0, mb1, mb2)
           )
\end{lstlisting}

\section{CPC bolts implementation code}\label{ap:cpc}

\subsection{Encoder}

\begin{lstlisting}
  from pl_bolts.models.self_supervised.cpc import AMDIMEncoder
\end{lstlisting}

\subsection{Transforms}

\begin{lstlisting}
from pl_bolts.models.self_supervised.cpc import CPCEvalTransformsSTL10

dataset = STL10(transforms=CPCEvalTransformsSTL10())
\end{lstlisting}

\subsection{Representation extraction}
\begin{lstlisting}
from pl_bolts.losses.self_supervised_learning import CPCTask

def cpc_task_forward(Z):
    losses = []

    context = context_cnn(Z)
    targets = target_cnn(Z)

    _, _, h, w = Z.shape

    # future prediction
    preds = pred_cnn(context)
    for steps_to_ignore in range(h - 1):
        for i in range(steps_to_ignore + 1, h):
            loss = compute_loss_h(targets, preds, i)
            if not torch.isnan(loss):
                losses.append(loss)

    loss = torch.stack(losses).sum()
    return loss
\end{lstlisting}

\subsection{Loss}

\begin{lstlisting}
def compute_loss_h(targets, preds, i):
    b, c, h, w = targets.shape

    # (b, c, h, w) -> (num_vectors, emb_dim)
    # every vector (c-dim) is a target
    targets = targets.permute(0, 2, 3, 1).contiguous().reshape([-1, c])

    # select the future (south) targets to predict
    # selects all of the ones south of the current source
    embed_scale = 0.1
    preds_i = preds[:, :, :-(i + 1), :] * embed_scale

    # (b, c, h, w) -> (b*w*h, c) (all features)
    # this ordering matches the targets
    preds_i = preds_i.permute(0, 2, 3, 1).reshape([-1, self.target_dim])

    # calculate the strength scores
    logits = torch.matmul(preds_i, targets.transpose(-1, -2))

    # generate the labels
    n = b * (h - i - 1) * w
    b1 = torch.arange(n) // ((h - i - 1) * w)
    c1 = torch.arange(n) % ((h - i - 1) * w)
    labels = b1 * h * w + (i + 1) * w + c1

    loss = cross_entropy(logits, labels)
    return loss
\end{lstlisting}

\subsection{Full CPC pseudocode}

\begin{lstlisting}
def training_step(self, batch, batch_num):
    x, y = batch
    
    # transform
    x = augmentations(x)
    x = patch_augmentation(x)

    # Latent features
    Z = encoder(x)

    # infoNCE loss
    nce_loss = self.contrastive_task(Z)
    loss = nce_loss
    return loss
\end{lstlisting}

\section{SimCLR bolts implementation code}\label{ap:simclr}

\subsection{Encoder}

\begin{lstlisting}
  from pl_bolts.models.self_supervised.resnets import resnet50_bn
\end{lstlisting}

\subsection{Transforms}

\begin{lstlisting}
from pl_bolts.models.self_supervised.cpc import SimCLREvalTransformsSTL10

dataset = STL10(transforms=SimCLREvalTransformsSTL10())
\end{lstlisting}

\subsection{Representation extraction}
\begin{lstlisting}
def training_step(self, batch, batch_idx):
    (img1, img2), y = batch

    # ENCODE
    # encode -> representations
    # (b, 3, 32, 32) -> (b, 2048, 2, 2)
    h1 = self.encoder(img1)
    h2 = self.encoder(img2)
\end{lstlisting}

\subsection{Loss}

\begin{lstlisting}
def nt_xent_loss(z1, z2, temperature):
    """
    Loss used in SimCLR
    """
    out = torch.cat([z1, z2], dim=0)
    n_samples = len(out)

    # Full similarity matrix
    cov = torch.mm(out, out.t().contiguous())
    sim = torch.exp(cov / temperature)

    # Negative similarity
    mask = ~torch.eye(n_samples, device=sim.device).bool()
    neg = sim.masked_select(mask).view(n_samples, -1).sum(dim=-1)

    # Positive similarity :
    pos = torch.exp(torch.sum(out_1 * out_2, dim=-1) / temperature)
    pos = torch.cat([pos, pos], dim=0)
    loss = -torch.log(pos / neg).mean()

    return loss

\end{lstlisting}

\subsection{Full SimCLR pseudocode}

\begin{lstlisting}
def training_step(self, batch, batch_idx):
    (img1, img2), y = batch

    # ENCODE
    # encode -> representations
    # (b, 3, 32, 32) -> (b, 2048, 2, 2)
    h1 = self.encoder(img1)
    h2 = self.encoder(img2)

    # the bolts resnets return a list of feature maps
    if isinstance(h1, list):
        h1 = h1[-1]
        h2 = h2[-1]

    # PROJECT
    # img -> E -> h -> || -> z
    # (b, 2048, 2, 2) -> (b, 128)
    z1 = self.projection(h1)
    z2 = self.projection(h2)

    loss = self.nt_xent_loss(z1, z2, self.hparams.loss_temperature)

    return loss
\end{lstlisting}

\end{document}